\documentclass{article}

 \usepackage[preprint]{neurips_2025}

\usepackage[utf8]{inputenc} 
\usepackage[T1]{fontenc}    
\usepackage{hyperref}       
\usepackage{url}            
\usepackage{graphicx}       
\usepackage{booktabs}       
\usepackage{array}          
\usepackage{wrapfig}        
\usepackage{caption}        
\usepackage{amsfonts}       
\usepackage{nicefrac}       
\usepackage{microtype}      
\usepackage[table]{xcolor}  
\usepackage{bbding}
\usepackage{multirow}
\usepackage{fontawesome5}
\definecolor{navy}{RGB}{30,74,117}

\newcommand{\ie}{\textit{i}.\textit{e}.}
\newcommand{\eg}{\textit{e}.\textit{g}.}
\newcommand{\etc}{\textit{etc}.}

\title{%
  \texorpdfstring{%
    \hspace*{-2.5em}
    \raisebox{-0.56\height}[0pt][0pt]{\includegraphics[height=1.2cm]{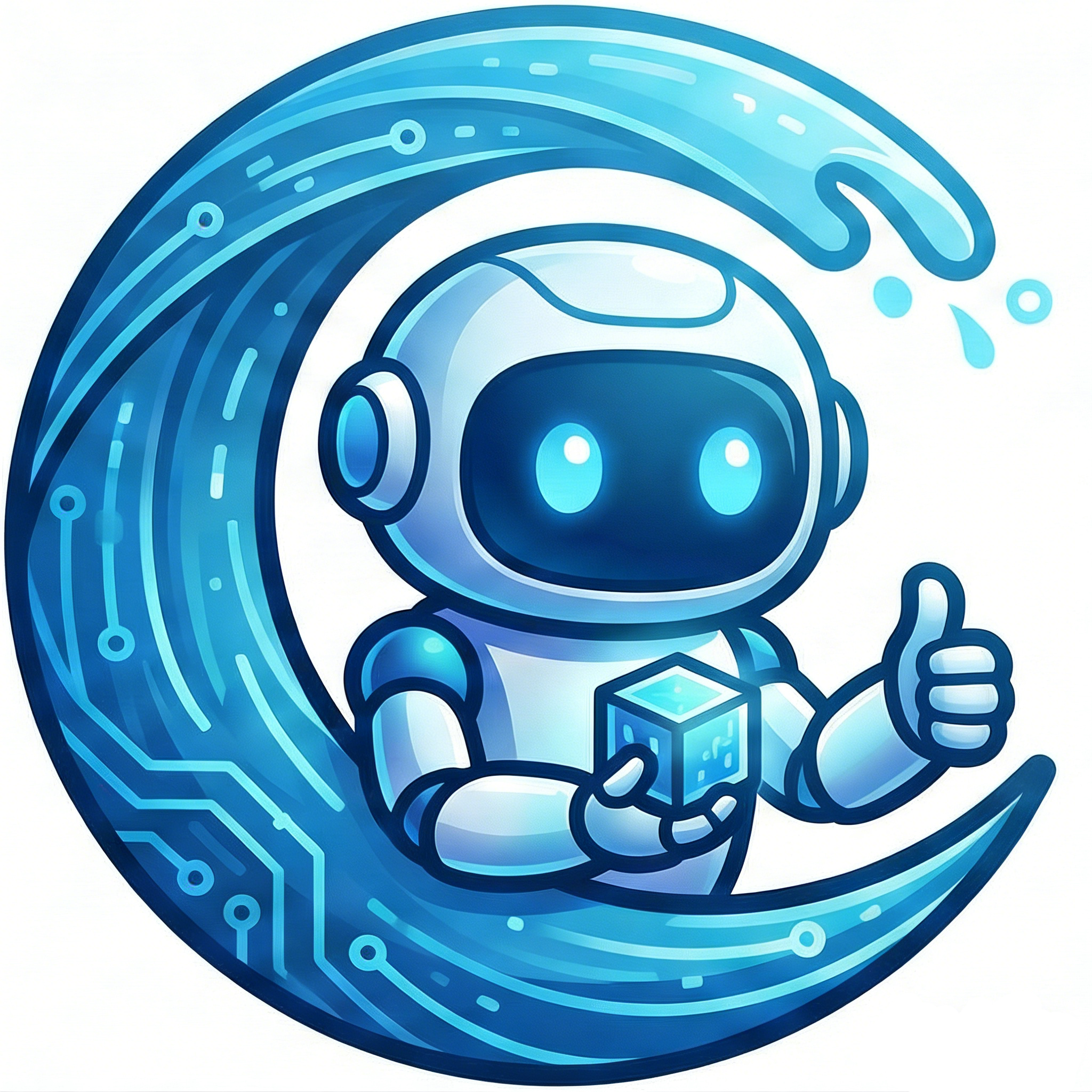}}
    \hspace{0.3em}%
  }{}%
  CoVe: Training Interactive Tool-Use Agents \\
  via Constraint-Guided Verification%
}

\author{%
  Jinpeng Chen$^{1}$\thanks{Equal contribution} \quad
  Cheng Gong$^{1}$\footnotemark[1] \quad
  Hanbo Li$^{2}$\footnotemark[1] \quad
  \textbf{Ziru Liu}$^{1}$ \quad
  \textbf{Zichen Tian}$^{2}$ \quad
  \textbf{Xinyu Fu}$^{1}$ \\
  \textbf{Shi Wu}$^{1}$ \quad
  \textbf{Chenyang Zhang}$^{1}$ \quad
  \textbf{Wu Zhang}$^{1}$ \quad
  \textbf{Suiyun Zhang}$^{1}$ \quad
  \textbf{Dandan Tu}$^{1}$\textsuperscript{\Envelope} \quad
  \textbf{Rui Liu}$^{1}$\textsuperscript{\Envelope} \\
  \\
  $^1$Huawei Research \qquad $^2$Independent Researcher \\
  \\
  \texttt{\{chen.jinpeng1, gong.cheng3, liu.rui2\}@huawei.com} \\
   \texttt{alexanderhanboli@gmail.com}
}

\begin{document}

\maketitle

\begin{abstract}
  Developing multi-turn interactive tool-use agents is challenging because real-world user needs are often complex and ambiguous, yet agents must execute deterministic actions to satisfy them. To address this gap, we introduce \textbf{CoVe} (\textbf{Co}nstraint-\textbf{Ve}rification), a post-training data synthesis framework designed for training interactive tool-use agents while ensuring both data complexity and correctness. CoVe begins by defining explicit task constraints, which serve a dual role: they guide the generation of complex trajectories and act as deterministic verifiers for assessing trajectory quality. This enables the creation of high-quality training trajectories for supervised fine-tuning (SFT) and the derivation of accurate reward signals for reinforcement learning (RL). Our evaluation on the challenging $\tau^2$-bench benchmark demonstrates the effectiveness of the framework. Notably, our compact \textbf{CoVe-4B} model achieves success rates of 43.0\% and 59.4\% in the Airline and Retail domains, respectively; its overall performance significantly outperforms strong baselines of similar scale and remains competitive with models up to $17\times$ its size. These results indicate that CoVe provides an effective and efficient pathway for synthesizing training data for state-of-the-art interactive tool-use agents. To support future research, we open-source our code, trained model, and the full set of 12K high-quality trajectories used for training.
\end{abstract}

\begin{center}
    \newcommand{\linkcolor}{navy}
    \renewcommand{\arraystretch}{1.5}
    \begin{tabular}{@{}c@{\hspace{0.7em}}l@{\hspace{0.7em}}l@{}}
        \textcolor{black}{\faRobot} & \textbf{Model} & \href{https://huggingface.co/Zichen1024/CoVe-4B}{\color{\linkcolor}https://huggingface.co/Zichen1024/CoVe-4B} \\
        \textcolor{black}{\faDatabase} & \textbf{Dataset} & \href{https://huggingface.co/datasets/Zichen1024/CoVe-12k}{\color{\linkcolor}https://huggingface.co/datasets/Zichen1024/CoVe-12k} \\
        \textcolor{black}{\faGlobe} & \textbf{Website} & \href{https://cove-agent.github.io}{\color{\linkcolor}https://cove-agent.github.io}
    \end{tabular}
\end{center}

\section{Introduction}
\label{sec:intro}

The rapid evolution of Large Language Models (LLMs) \citep{achiam2023gpt,bai2023qwen} has catalyzed a paradigm shift from text generation to task automation \citep{team2025kimi,liu2025deepseek}, enabling agents to leverage external tools for solving real-world problems. While current models excel in simple single-step tasks or non-interactive multi-tool execution, developing agents capable of engaging in multi-turn interactions interleaved with appropriate tool use remains a significant challenge \citep{yao2024tau, barres2025tau, patil2025berkeley}. This difficulty stems from a fundamental misalignment between human communication and machine execution: real-world user needs are often complex, implicit, and ambiguous, whereas underlying tools (APIs) strictly require deterministic actions and precise arguments to function. Consequently, an effective agent must not only comprehend vague instructions but also clarify user intent through multi-turn interactions before translating them into executable commands.

To acquire these capabilities, high-quality training data is paramount. However, acquiring such data presents a significant bottleneck. Human annotation is prohibitively expensive and difficult to scale, rendering large-scale training infeasible. As an alternative, existing synthesis methods rely on LLMs to generate user–agent interaction trajectories with tool calls, and also use LLMs to verify whether the trajectories are correct \citep{prabhakar2025apigen,li2025simulating,xu2026unlocking}. However, due to the inherent uncontrollability of LLMs, these approaches often fail to guarantee the solvability of queries or the absolute correctness of trajectory verification. Moreover, constrained by their own capabilities, LLMs tend to generate tasks with limited tool invocations and dialogue turns, thereby inhibiting the emergence of complex, challenging samples. Consequently, there is an urgent need for a post-training data synthesis framework that can simultaneously ensure both data complexity and trajectory correctness.

\begin{figure}
    \centering
    \includegraphics[width=\textwidth]{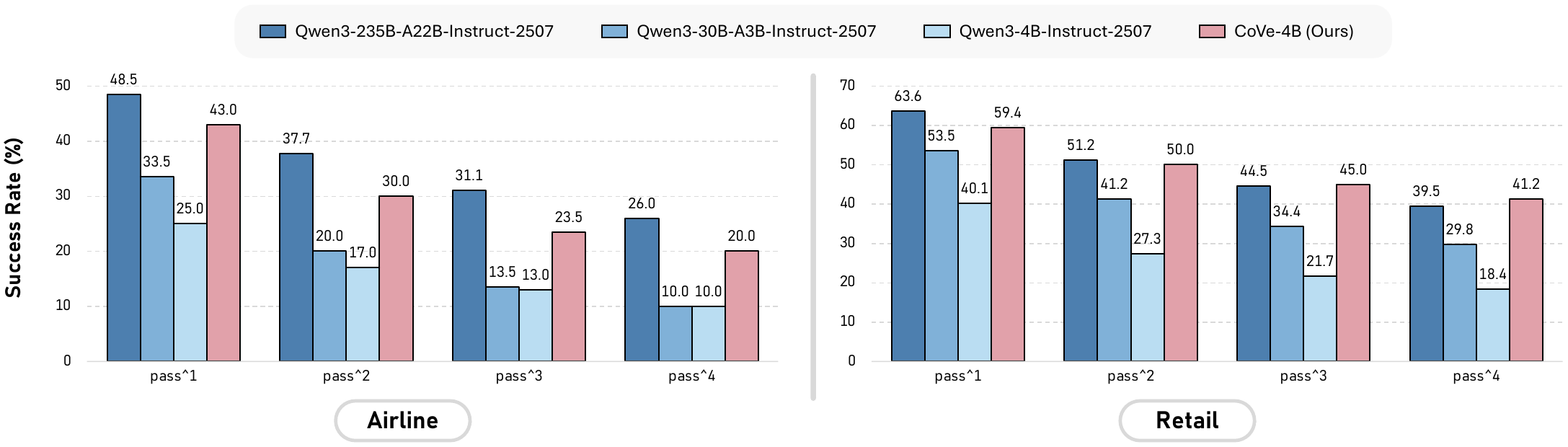}
    \caption{Performance evaluation on the $\tau^2$-bench Airline and Retail domains. We report success rates across $pass^1$ to $pass^4$ metrics to assess both peak performance and stability.}
    \label{fig:first_histogram}
\end{figure}

To bridge this gap, we introduce \textbf{CoVe} (\textbf{Co}nstraint-\textbf{Ve}rification), a robust framework designed to synthesize high-quality interactive tool-use data. CoVe begins by sampling explicit task constraints, \ie, a set of deterministic goals the agent must achieve, such as specific departure cities and hotel tiers in travel planning, or items to return and address updates in e-commerce customer service. Subsequently, we employ a constraint fuzzification strategy to mimic real-world ambiguity; for instance, an order might be referenced by its contents (\eg, "the order with a blue shirt and a pair of leather shoes") rather than a precise Order ID. These fuzzified constraints guide a User Simulator LLM to engage in multi-turn interactions with the agent, progressively revealing needs. Upon conversation completion, the original constraints serve as a deterministic checklist, allowing us to rigorously verify via rule-based matching whether the agent's tool invocations have satisfied every requirement.

This mechanism enables automated, large-scale data synthesis with guaranteed quality. Since tasks are derived from pre-defined constraints, solvability is inherent, and the verification of tool execution is exact, eliminating the hallucinations common in LLM-based evaluation. Moreover, by composing multiple constraints combined with fuzzified expressions, CoVe fosters the emergence of highly complex tasks. This framework seamlessly supports both Supervised Fine-Tuning (SFT) and Reinforcement Learning (RL). For SFT, a teacher model acts as the agent to generate trajectories, and only those verified as fully correct are retained for training. For RL, the same deterministic verifier provides precise reward signals to the policy agent model, enabling it to learn from exploration within a self-improving loop.

We evaluate the effectiveness of CoVe on the challenging Airline and Retail domains of the $\tau^2$-bench. As shown in Figure \ref{fig:first_histogram}, empirical results demonstrate that our framework significantly enhances agent performance: our compact CoVe-4B model achieves an overall success rate of 51.2\%, outperforming strong baselines with similar parameter scales and proving competitive with open-source models up to 70B in size. These results validate CoVe as a robust and effective pathway for synthesizing training data for state-of-the-art interactive agents. Our main contributions are summarized as follows:

\begin{itemize}
    \item We propose CoVe, a post-training data synthesis framework that generates multi-turn interactive tool-use trajectories and performs deterministic verification based on explicit constraints, ensuring both complexity and correctness.
    \item We demonstrate that data synthesized by CoVe effectively supports both SFT and RL. Our CoVe-4B model achieves strong performance despite its smaller scale, surpassing state-of-the-art models of similar or moderately larger sizes, and performing competitively against models with several times more parameters.
    \item To facilitate future research, we open-source our code, the trained model, and the full set of 12K high-quality interactive tool-use trajectories used to train the model.
\end{itemize}
\section{Related Work}
\label{sec:related_work}

\subsection{Tool-Use Agents}

Foundational research in tool-use agents, such as TALM \citep{parisi2022talm}, Toolformer \citep{schick2023toolformer}, and ReAct \citep{yao2022react}, has demonstrated the potential of LLMs to autonomously interact with external APIs. As model capabilities have evolved, these agents have been deployed in increasingly complex domains, including deep research \citep{huang2025deep,team2025tongyi}, code generation \citep{zhang2024codeagent,yang2024swe}, GUI navigation \citep{nguyen2025gui,wang2024mobile,guan2025kg}, \etc. However, these works primarily focus on scenarios initiated by a self-contained query, where the initial instruction provides all requirements, and the agent executes tasks without further user interaction. Recently, inspired by pioneering benchmarks \citep{yao2024tau,barres2025tau,patil2025berkeley}, research on multi-turn interactive tool-use agents has begun to emerge. In this setting, agents are initially presented with incomplete or ambiguous user needs, necessitating an interleaved process of user clarification and tool execution to progressively resolve the task.

\subsection{Multi-turn Interactive Tool-Use Data Synthesis}

High-quality data is paramount for advancing multi-turn interactive tool-use agents; however, its acquisition is prohibitively expensive and difficult to scale. Consequently, various automated synthesis frameworks have emerged to address this bottleneck. APIGen-MT \citep{prabhakar2025apigen} adopts a "blueprint-to-dialogue" pipeline to synthesize multi-turn trajectories with quality verification. Simia \citep{li2025simulating} leverages LLMs to simulate environment and user feedback, facilitating scalable training for both SFT and RL. GEM \citep{xu2026unlocking} takes a different approach by mining implicit multi-step processes from large-scale text corpora and transforming them into executable trajectories. MUA-RL \citep{zhao2025mua} integrates dynamic user simulation within reinforcement learning to optimize policies for realistic interaction scenarios. In contrast to these approaches, we leverage explicit task constraints as the genesis of data generation. This strategy not only effectively amplifies task complexity but also guarantees that execution outcomes can be verified with deterministic accuracy.
\section{Problem Formulation}
\label{sec:problem_def}

We formalize the multi-turn interaction between a tool-use agent and a user as a simplified Partially Observable Markov Decision Process (POMDP), defined by the tuple $\mathcal{M} = (\mathcal{S}, \mathcal{A}, \mathcal{O}, \mathcal{T})$.

\begin{itemize}
    \item $\mathcal{S}$ represents the latent state space, which consists of the user's hidden intent $u$ and the external environment state $e$.
    \item $\mathcal{A} = \mathcal{A}_{\mathrm{tool}} \cup \mathcal{A}_{\mathrm{resp}}$ denotes the action space. The agent can either execute a structured tool call $a \in \mathcal{A}_{\mathrm{tool}}$ to query/modify the environment, or generate a natural language response $a \in \mathcal{A}_{\mathrm{resp}}$ to communicate with the user.
    \item $\mathcal{O}$ is the observation space. At each time step $t$, the observation $o_t$ can be either a user instruction $o^{\mathrm{user}}$ or the execution result $o^{\mathrm{env}}$ returned by a tool call.
    \item $\mathcal{T}(s_{t+1} \mid s_t, a_t)$ is the transition function governing how the environment state and user intent evolve based on the agent's actions.
\end{itemize}

An interaction session unfolds over multiple turns, forming a trajectory $\tau = [o_1, a_1, \dots, o_T, a_T]$. At each step $t$, the agent decides its action $a_t$ based on the interaction history $h_t = (o_1, a_1, \dots, o_t)$. The objective is to generate a trajectory $\tau$ that successfully fulfills the user's latent intent $u$ while adhering to the constraints of the external environment $e$.
\section{CoVe: Constraint-Guided Trajectory Generation and Deterministic Verification}
\label{sec:cove}

\begin{figure}
    \centering
    \includegraphics[width=\textwidth]{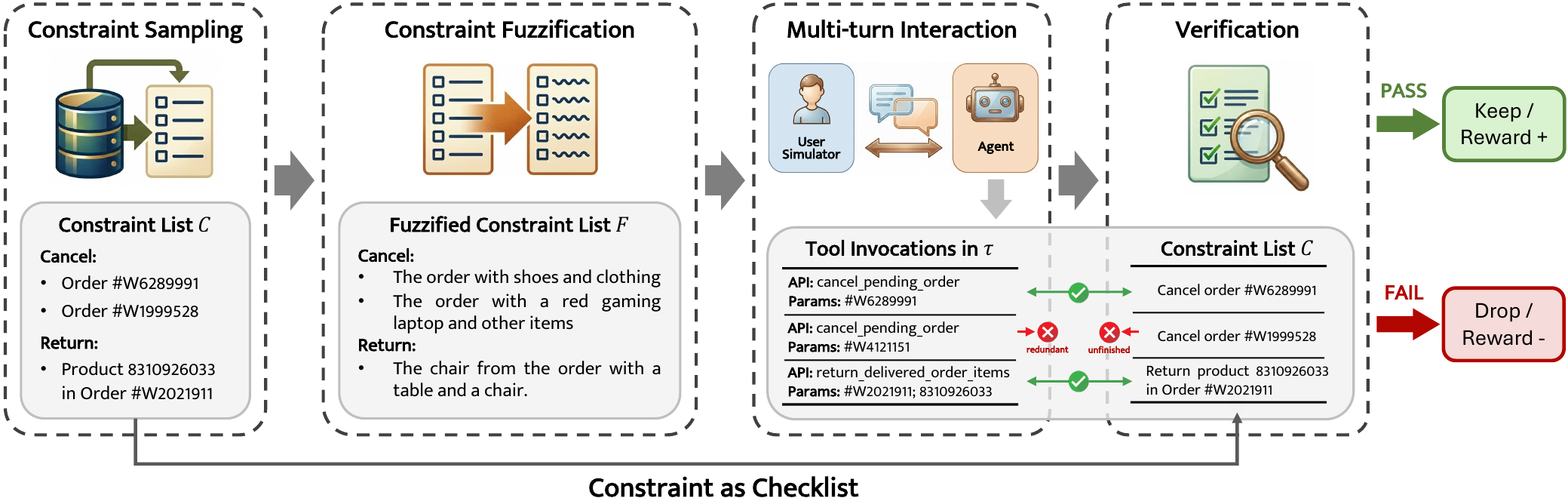}
    \caption{The CoVe framework. Explicit constraints are sampled and fuzzified to guide a User Simulator LLM in generating ambiguous, realistic queries. Upon conversation completion, the original deterministic constraints act as a checklist to automatically verify the agent's tool invocations.}
    \label{fig:pipeline}
\end{figure}

\subsection{Overview}

Figure \ref{fig:pipeline} illustrates the overall pipeline for trajectory generation and verification within the CoVe framework, along with an example from the Retail domain environment of $\tau^2$-bench \citep{barres2025tau}. The process begins with explicit constraint sampling, where a set of deterministic constraints $C = \{c_1, c_2, \dots, c_n\}$ is sampled from the sandbox databases with available tools to define the overall goals of a task. Subsequently, these explicit constraints (or their constituent elements) are transformed into a fuzzified form $F = \{f_1, f_2, \dots, f_m\}$ to mimic the ambiguity inherent in real-world user requests. Guided by $F$, a User Simulator LLM engages in multi-turn interactions with the Agent, yielding a trajectory $\tau$ comprising the conversation history and tool invocations. Finally, the process concludes with rule-based verification, where the original constraints $C$ serve as a ground-truth checklist. A deterministic verification function $V(\tau, C)$ rigorously evaluates whether the agent’s tool executions in $\tau$ satisfy all conditions specified in $C$ and are free of redundancy.

\subsection{Constraint Sampling and Fuzzification}
\label{sec:constraint_sampling_fuzzification}

The foundation of CoVe lies in constructing task goals that are both inherently solvable and realistically ambiguous. This process unfolds in two phases: \textit{Constraint Sampling} to establish the ground truth, and \textit{Constraint Fuzzification} to simulate natural user expressions.

\paragraph{Constraint Sampling.}
To ensure that every generated task is executable and free from hallucinations, we ground the generation process in a deterministic pre-task environment state, defined by the entries in the sandbox database with a list of available tools. From this state, we randomly sample a set of constraints $C = \{c_1, c_2, \dots, c_n\}$. As illustrated in Figure \ref{fig:pipeline}, $C$ defines the comprehensive requirements for the overall task. For instance, in the Retail domain of $\tau^2$-bench, a constraint $c_i \in C$ might specify the need to cancel a particular order or return a specific item. 

Since $C$ is derived directly from existing database records, the solvability of the task is logically guaranteed. During this sampling phase, we utilize precise identifiers (\eg, Order IDs, Product IDs) to refer to the elements involved in the constraints, as these identifiers are independently sufficient to uniquely identify specific elements in the environment.

\paragraph{Constraint Fuzzification.}
Directly exposing the explicit IDs in $C$ to the agent would reduce the problem to trivial slot-filling. Moreover, real-world users rarely recall precise identifiers; instead, they typically refer to objects by their composition, attributes, or context. To bridge this gap, we employ a fuzzification strategy that maps deterministic IDs to more ambiguous descriptions while preserving logical uniqueness.

Taking the Retail domain of the $\tau^2$-bench as a representative example, we design specific fuzzification strategies for five key elements based on domain policies and tool capabilities:

\begin{itemize}
    \item \textbf{User ID:} Fuzzified into an email address or a combination of the user's name and zip code. According to the domain policy, either representation uniquely identifies a specific user.
    \item \textbf{Order ID:} Replaced by a random subset of items contained in the order. We verify that among the user's order list, the target order is the \textit{only} one containing this specific combination of items.
    \item \textbf{Item ID:} Described by the item/product name, or a combination of the product and specific attributes (randomly sampled options). Similarly, we verify that this description uniquely pinpoints the item within the given context.
    \item \textbf{Payment ID:} Fuzzified into the payment type (e.g., Credit Card, Gift Card, PayPal). For credit cards, we may randomly include the card brand or the last four digits of the card number. In all cases, we ensure that the fuzzified description can uniquely identify the payment method.
    \item \textbf{Address:} If the target address matches one used in previous orders or the user's default address, it is described as ``an address used in other orders'' or ``default user address'' paired with the city name, provided this combination remains unique.
\end{itemize}

The processed constraints result in a set of fuzzified instructions $F = \{f_1, f_2, \dots, f_m\}$, which are then fed into the User Simulator LLM for trajectory generation. This ensures that the simulated user interaction mimics human ambiguity—compelling the agent to reason or query the database—while strictly maintaining a deterministic mapping back to the original constraints $C$.

\subsection{Trajectory Generation and Verification}
\label{sec:trajectory_generation_verification}

With the fuzzified instructions $F$ prepared, the subsequent stage of CoVe focuses on synthesizing high-quality interaction data and ensuring its validity. This workflow proceeds in two steps: \textit{Trajectory Generation}, where a User Simulator LLM engages the Agent in multi-turn dialogues, and \textit{Trajectory Verification}, which rigorously evaluates the resulting execution history against the deterministic ground-truth constraints $C$.

\begin{figure}
    \centering
    \includegraphics[width=\textwidth]{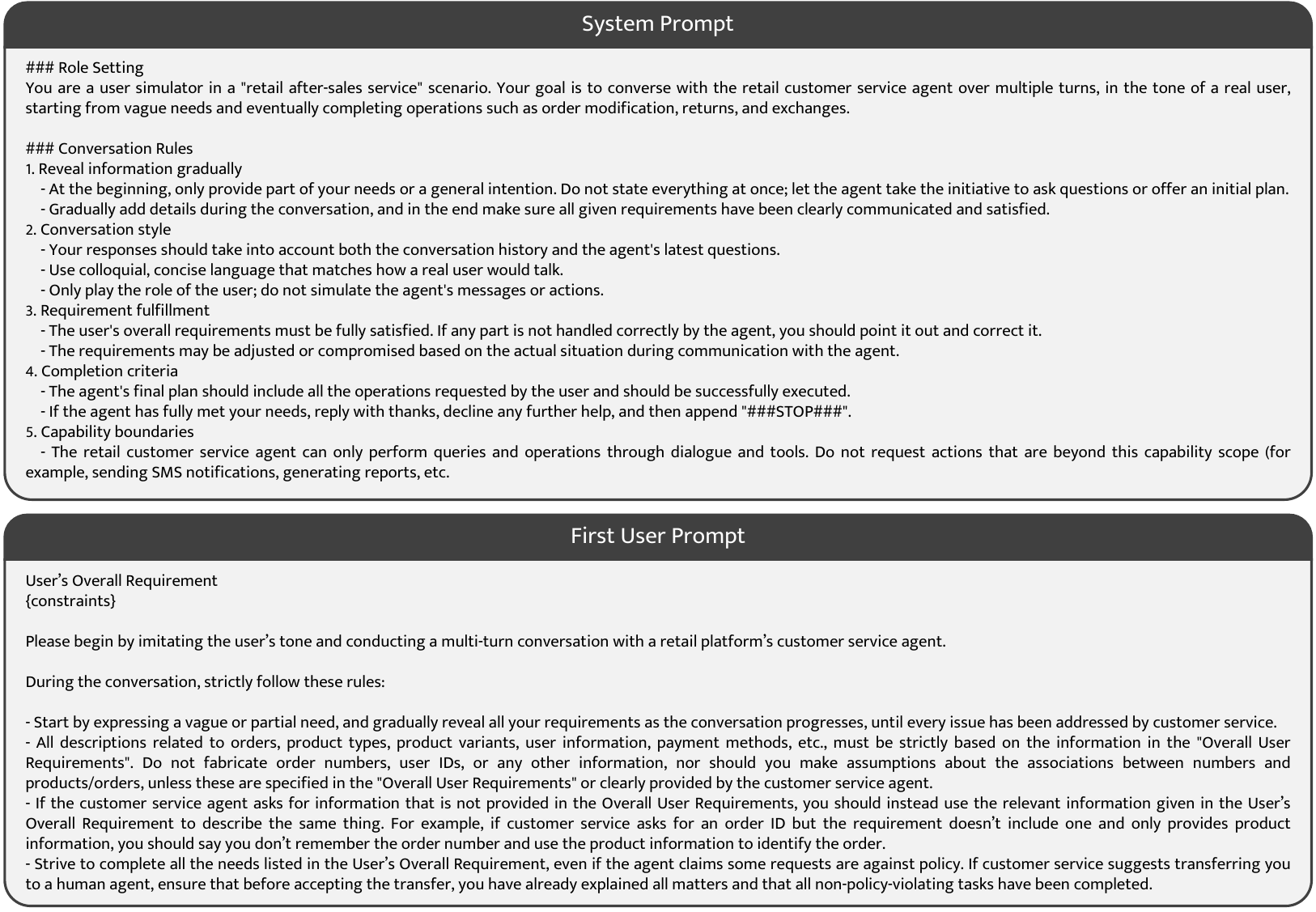}
    \caption{The system prompt (top) and the first user prompt (bottom) for the User Simulator. After receiving the first user prompt, the simulator begins issuing task requests to the Agent. In subsequent dialogue turns, the Agent's response serves as the user message for the User Simulator.}
    \label{fig:pipeline}
\end{figure}

\paragraph{Trajectory Generation.}
We employ a User Simulator LLM to assume the role of a human user, engaging in multi-turn interactions with the Agent. The instructions $F$ are embedded into the prompt given to the simulator. Crucially, the simulator is instructed to reveal the requirements within $F$ progressively and naturally, rather than disclosing all constraints at the outset. This design increases task difficulty and better mimics real-world communication dynamics. 

As the User Simulator gradually unfolds its needs, the Agent is tasked with clarifying ambiguities and synchronizing information through dialogue, along with invoking tools to query data or resolve requests. Following the protocol in \citep{barres2025tau}, we require the User Simulator to output a special termination marker (\eg, \texttt{\#\#\#STOP\#\#\#}) when it deems the task complete. This marker triggers the end of the session, and the Agent's entire context history is recorded as the trajectory $\tau$. The system prompt and first user prompt for the user simulator are provided in Figure \ref{fig:pipeline}.

\paragraph{Trajectory Verification.}
After generation, each trajectory $\tau$ is subjected to quality verification. Unlike prior methods that rely on LLM-based evaluation \citep{prabhakar2025apigen,li2025simulating}, CoVe leverages the explicit constraint set $C$ as a deterministic ground-truth checklist, thereby avoiding unreliable judgments caused by hallucinated assessments from LLMs.

Specifically, we design a rule-based verifier $V(\tau, C)$ that parses the tool invocation records in $\tau$ and checks them against each constraint $c_i \in C$. Importantly, the verifier evaluates whether the required outcome is achieved, rather than enforcing a predefined sequence of actions. For example, if a constraint specifies \textit{cancel order \#W6289991}, the verifier accepts any tool call or combination of calls in $\tau$ that leads to the cancellation of that order. This may include a direct cancellation action or canceling all items associated with the order, provided that these operations are considered equivalent under the relevant domain policy. A constraint is regarded as satisfied if any valid execution path results in the intended state. In addition, the verifier identifies and penalizes redundant operations that do not correspond to any constraint in $C$. Finally, we compute a composite score based on two factors: the constraint satisfaction rate (higher is better) and the number of redundant actions (lower is better).



\subsection{Post-Training with CoVe}
\label{sec:training_with_cove}

The trajectory generation and verification mechanism provided by CoVe naturally support two mainstream post-training paradigms: SFT and RL.

For \textbf{SFT}, we employ a teacher model as the Agent to interact with the User Simulator, generating a large number of candidate trajectories. Upon completion of each trajectory $\tau$, the verifier computes a final score $S_{\tau} = V(\tau, C)$ based on the constraint satisfaction and redundancy metrics defined in Section \ref{sec:trajectory_generation_verification}. We retain only those trajectories that achieve the maximum score of $1$, \ie, those that satisfy all constraints in the set $C$ without any redundant operations. This process yields a clean dataset containing exclusively correct execution paths, which is then used to fine-tune the target student model.

For \textbf{RL}, CoVe serves both as a robust training environment in which the student agent interacts with the User Simulator to explore solution paths and as a reward provider that assigns rewards based on the verification of the interaction trajectory. After a trajectory $\tau$ is completed, the final score $S_{\tau}$ produced by the verifier is directly fed back to the agent to guide policy updates. In this manner, the agent gradually learns to generate higher-quality execution paths that satisfy all constraints.



\section{Experiment}
\label{sec:experiment}

\subsection{Experimental Setup}

\paragraph{Implementation Details.}
We conduct all model training using the VeRL framework \citep{sheng2025hybridflow}, utilizing Qwen3-4B-Instruct-2507 \citep{yang2025qwen3} as the base model. The experiments are performed on two computational nodes, each equipped with eight 80GB GPUs. For SFT, we employ the AdamW optimizer with a learning rate of $1\times10^{-6}$ and a global batch size of 128. For RL, we adopt the Group Relative Policy Optimization (GRPO) algorithm \citep{shao2024deepseekmath}, maintaining the learning rate at $1\times10^{-6}$ with a training batch size of 64. We utilize SGLang \citep{zheng2024sglang} as the rollout engine, with a sampling temperature of 1.0. For advantage estimation, we sample 16 rollouts per prompt, setting the maximum number of user turns and assistant turns to 32 and 40, respectively. Unless otherwise specified, our primary model \textbf{CoVe-4B} refers to the version trained exclusively via SFT.

\paragraph{Benchmarks and Metrics.}
To evaluate model performance in realistic and complex interactive scenarios, we evaluate our CoVe-trained models on the Airline and Retail domains of $\tau^2$-bench \citep{barres2025tau}. Following the default $\tau^2$-bench evaluation protocol, we disable the \textit{think} tool. For the main evaluation, we compare our primary model, CoVe-4B, against representative proprietary state-of-the-art models as well as open-source baselines \citep{yang2025qwen3,prabhakar2025apigen,li2025simulating,xu2026unlocking} across various parameter scales.

Regarding evaluation metrics, we strictly follow the protocol defined in \citep{barres2025tau} and report results using the $pass^k$ metrics, specifically ranging from $pass^1$ to $pass^4$. These metrics quantify the probability of the model successfully completing a task in all $k$ consecutive independent runs. This rigorous standard is designed to assess the agent's stability and consistency in execution, differentiating it from single-attempt performance.

\paragraph{Environment and Data.}
To avoid data contamination, the sandbox database we sample from is generated using LLMs and hand-crafted rules. It matches the style of the official $\tau^2$-bench database but differs in its specific content. For SFT data generation, we use multiple user-simulator LLMs such as Qwen3-235B-A22B-Instruct-2507 and Gemini-3-Pro to increase data diversity and reduce biases that may arise from relying on a single simulator. The teacher Agent model used for SFT data production is Qwen3-235B-A22B-Instruct-2507. During the RL stage, due to cost constraints, we use only Qwen3-235B-A22B-Instruct-2507 as the User Simulator LLM.

\begin{table}[t]
  \centering
  \caption{Main results evaluating proprietary and open-source models on the Airline and Retail domains of $\tau^2$-bench. \textbf{Bold} and \underline{underline} denote the best and second-best results within each group.}
  \scriptsize
  \renewcommand{\arraystretch}{1.3}
  \newcommand{\passonewidth}{0.44cm}
  \newcommand{\passtwowidth}{0.49cm}
  \newcommand{\passthreewidth}{0.47cm}
  \newcommand{\passfourwidth}{0.56cm}
  \begin{tabular}{l|>{\centering\arraybackslash}p{\passonewidth}>{\centering\arraybackslash}p{\passtwowidth}>{\centering\arraybackslash}p{\passthreewidth}>{\centering\arraybackslash}p{\passfourwidth}|>{\centering\arraybackslash}p{\passonewidth}>{\centering\arraybackslash}p{\passtwowidth}>{\centering\arraybackslash}p{\passthreewidth}>{\centering\arraybackslash}p{\passfourwidth}|>{\centering\arraybackslash}p{\passonewidth}>{\centering\arraybackslash}p{\passtwowidth}>{\centering\arraybackslash}p{\passthreewidth}>{\centering\arraybackslash}p{\passfourwidth}}
    \toprule
    \multirow{2}{*}{\textbf{Model}} &
    \multicolumn{4}{c}{\textbf{Airline}} &
    \multicolumn{4}{c}{\textbf{Retail}} &
    \multicolumn{4}{c}{\textbf{Average}} \\
    \cmidrule(lr){2-5}
    \cmidrule(lr){6-9}
    \cmidrule(lr){10-13}
    & $pass^{1}$ & $pass^{2}$ & $pass^{3}$ & $pass^{4}$ &
      $pass^{1}$ & $pass^{2}$ & $pass^{3}$ & $pass^{4}$ &
      $pass^{1}$ & $pass^{2}$ & $pass^{3}$ & $pass^{4}$ \\
    \midrule
    \multicolumn{13}{c}{\textit{Proprietary Models}} \\
    \midrule
    GPT-5 & \textbf{62.5\%} & \textbf{55.3\%} & \textbf{51.0\%} & \textbf{48.0\%} & \textbf{81.6\%} & \textbf{71.8\%} & \textbf{64.7\%} & \textbf{58.8\%} & \textbf{72.1\%} & \textbf{63.6\%} & \textbf{57.9\%} & \textbf{53.4\%} \\
    GPT-4o & \underline{47.6\%} & \underline{35.7\%} & \underline{30.0\%} & \underline{26.5\%} & \underline{64.0\%} & \underline{50.8\%} & \underline{43.0\%} & \underline{37.7\%} & \underline{55.8\%} & \underline{43.3\%} & \underline{36.5\%} & \underline{32.1\%} \\
    \midrule
    \multicolumn{13}{c}{\textit{Open-Source Models ($\geq$70B)}} \\
    \midrule
    Qwen3-235B-A22B-Inst. & \textbf{48.5\%} & \textbf{37.7\%} & \textbf{31.0\%} & \textbf{26.0\%} & \textbf{63.6\%} & \textbf{51.2\%} & \textbf{44.5\%} & \textbf{39.5\%} & \textbf{56.1\%} & \textbf{44.5\%} & \textbf{37.8\%} & \textbf{32.8\%} \\
    xLAM-2-70b-fc-r & \underline{43.3\%} & \underline{31.2\%} & \underline{25.2\%} & \underline{20.7\%} & \underline{59.7\%} & \underline{44.1\%} & \underline{35.8\%} & \underline{30.3\%} & \underline{51.5\%} & \underline{37.6\%} & \underline{30.5\%} & \underline{25.5\%} \\
    \midrule
    \multicolumn{13}{c}{\textit{Open-Source Models ($\sim$30B)}} \\
    \midrule
    Simia-Tau-32B & \textbf{48.5\%} & \textbf{36.3\%} & \textbf{32.0\%} & \textbf{30.0\%} & \textbf{62.5\%} & \textbf{50.1\%} & \textbf{41.9\%} & \textbf{36.0\%} & \textbf{55.5\%} & \textbf{43.2\%} & \textbf{37.0\%} & \textbf{33.0\%} \\
    xLAM-2-32b-fc-r & \underline{47.0\%} & \textbf{36.3\%} & \underline{31.5\%} & \underline{28.0\%} & 52.1\% & 37.1\% & 29.2\% & 24.6\% & \underline{49.5\%} & \underline{36.7\%} & \underline{30.3\%} & \underline{26.3\%} \\
    GEM-32B & 35.5\% & - & - & - & \underline{55.5\%} & - & - & - & 45.5\% & - & - & - \\
    Qwen3-30B-A3B-Inst. & 33.5\% & 20.0\% & 13.5\% & 10.0\% & 53.5\% & \underline{41.2\%} & \underline{34.4\%} & \underline{29.8\%} & 43.5\% & 30.6\% & 24.0\% & 19.9\% \\
    \midrule
    \multicolumn{13}{c}{\textit{Open-Source Models ($\leq$8B)}} \\
    \midrule
    Simia-Tau-8B & 42.5\% & \underline{33.0\%} & \textbf{29.0\%} & \textbf{26.0\%} & \underline{52.4\%} & 39.6\% & 32.7\% & 28.1\% & 47.5\% & 36.3\% & 30.9\% & 27.1\% \\
    Simia-Tau-RL-8B & \textbf{43.0\%} & \textbf{34.0\%} & \textbf{29.0\%} & \textbf{26.0\%} & \underline{52.4\%} & \underline{41.1\%} & \underline{34.9\%} & \underline{30.7\%} & \underline{47.7\%} & \underline{37.6\%} & \underline{32.0\%} & \underline{28.4\%} \\
    xLAM-2-8b-fc-r & 33.5\% & 19.2\% & 12.5\% & 9.0\% & 48.9\% & 34.5\% & 27.5\% & 22.5\% & 41.2\% & 26.8\% & 20.0\% & 15.9\% \\
    GEM-8B & 22.0\% & - & - & - & 44.5\% & - & - & - & 33.3\% & - & - & - \\
    Qwen3-4B-Inst. & 25.0\% & 17.0\% & 13.0\% & 10.0\% & 40.1\% & 27.3\% & 21.7\% & 18.4\% & 32.6\% & 22.2\% & 17.4\% & 14.2\% \\
    Simia-Tau-3B & 36.0\% & - & - & - & 32.0\% & - & - & - & 34.0\% & - & - & - \\
    xLAM-2-3b-fc-r & 24.0\% & 12.7\% & 7.5\% & 4.0\% & 15.1\% & 8.8\% & 6.4\% & 5.3\% & 19.6\% & 10.8\% & 7.0\% & 4.7\% \\
    \rowcolor{blue!15}
    CoVe-4B (Ours) & \textbf{43.0\%} & 30.0\% & 23.5\% & 20.0\% & \textbf{59.4\%} & \textbf{50.0\%} & \textbf{45.0\%} & \textbf{41.2\%} & \textbf{51.2\%} & \textbf{40.0\%} & \textbf{34.3\%} & \textbf{30.6\%} \\
    \bottomrule
  \end{tabular}
  \label{tab:main_results}
\end{table}

\subsection{Main Results}

Table \ref{tab:main_results} presents the comprehensive evaluation results of our CoVe-4B model against proprietary models and various open-source baselines categorized by parameter scales. 

Within the $\leq$8B parameter group, CoVe-4B demonstrates highly competitive performance. On average, it achieves a $pass^1$ score of 51.2\%, securing the top position in its class over strong baselines like Simia-Tau-RL-8B (47.7\%) and xLAM-2-8b-fc-r (41.2\%). Across both evaluated domains, the model demonstrates consistently superior performance, achieving the highest $pass^1$ scores in each leaderboard. Its advantage is especially pronounced in the Retail domain, where it maintains leading results from $pass^1$ through $pass^4$. Furthermore, compared to its base model, Qwen3-4B-Instruct-2507, our training pipeline yields a striking absolute improvement of +18.6\% in average $pass^1$ (from 32.6\% to 51.2\%).

The most compelling finding is CoVe-4B's ability to match or surpass several models with significantly larger parameter counts. In terms of average $pass^1$, CoVe-4B (51.2\%) successfully outperforms a portion of mid-scale ($\sim$30B) models, such as xLAM-2-32b-fc-r (49.5\%) and the Qwen3-30B-Inst. base model (43.5\%). More impressively, it closely approaches the performance of the massive xLAM-2-70b-fc-r (51.5\%), effectively bridging the capability gap with a model nearly 17$\times$ its size. Even when compared against colossal open-source and proprietary models like Qwen3-235B-A22B-Instruct-2507 and GPT-4o, the performance gap remains remarkably narrow at just 4.9\% and 4.6\%, respectively. This cross-scale superiority demonstrates the overall effectiveness of the CoVe framework, indicating that synthesizing realistic, complex task constraints and strictly filtering for high-quality trajectories can successfully compensate for the limited capacity of smaller LLMs.

\subsection{Ablation and Analysis}

\begin{wrapfigure}{r}{0.46\linewidth}
  \begin{minipage}{1\linewidth}
    \centering
    \scriptsize
    \captionof{table}{Performance comparison ($pass^1$) for the Qwen3-4B-Instruct-2507\ model fine-tuned on different SFT datasets. \textbf{Bold} and \underline{underline} denote the best and second-best results.}
    \renewcommand{\arraystretch}{1.3}
    \begin{tabular}{l|ccc}
      \toprule
      \textbf{Training Set} &
      \textbf{Airline} &
      \textbf{Retail} &
      \textbf{Average} \\
      \midrule
      None & 25.0\% & 40.1\% & 32.6\% \\
      \midrule
      APIGen-MT-5K & 34.0\% & 49.3\% & 41.7\% \\
      Simia-5K & 33.0\% & 46.3\% & 39.7\% \\
      Simia-90K & \underline{42.0\%} & 46.5\% & 44.3\% \\
      \midrule
      CoVe-5K (Ours) & 38.8\% & \underline{50.5\%} & \underline{44.7\%} \\
      CoVe-12K (Ours) & \textbf{43.0\%} & \textbf{59.4\%} & \textbf{51.2\%} \\
      \bottomrule
    \end{tabular}
    \label{tab:sft_data}
  \end{minipage}
  \end{wrapfigure}

\paragraph{Impact of Data Quality and Scaling.}
To isolate the effectiveness of our data synthesis pipeline, we fine-tune the same base model (Qwen3-4B-Instruct-2507) using various SFT datasets and evaluate their performance, as detailed in Table \ref{tab:sft_data}. When controlling the dataset size to 5K trajectories, CoVe-5K achieves an average $pass^1$ of 44.7\%, significantly outperforming APIGen-MT-5K (41.7\%) and Simia-5K (39.7\%). This demonstrates that the deterministic constraint sampling and rigorous verification mechanism within CoVe yield cleaner, more effective training signals. The most striking observation emerges when comparing CoVe-5K with the massive Simia-90K dataset: despite utilizing merely $\sim$5.5\% of the data volume, CoVe-5K manages to achieve a slightly higher average score (44.7\% vs. 44.3\%). This proves that carefully curated, zero-redundancy trajectories are vastly superior to simply scaling up noisier interaction data. Finally, expanding our dataset to CoVe-12K yields a substantial performance leap across both domains, pushing the average $pass^1$ to an impressive 51.2\% and indicating that scaling up high-quality, CoVe-verified trajectories consistently unlocks further model potential.

\paragraph{Comparison of Training Paradigms.}
To evaluate the impact of different training paradigms within the CoVe framework, we compare the performance of our base model under pure SFT, pure RL, and sequential training (SFT+RL), as shown in the left part of Table \ref{tab:rl_sft_comparison_and_generation_success_rate}. Both pure SFT (51.2\%) and pure RL (40.7\%) yield substantial improvements over the Qwen3-4B-Instruct-2507 baseline (32.6\%), confirming that our deterministic verification mechanism is highly effective both as an offline data filter and as an online reward signal. Notably, the pure SFT approach achieves the highest overall performance. While a common post-training pipeline involves sequential training, we observe a performance degradation when applying RL on top of the already strong SFT model (dropping to 46.9\%). We attribute this counter-intuitive result to an \textit{environment bottleneck} caused by practical resource constraints. During SFT data generation, trajectories are synthesized using a diverse ensemble of highly capable models like Gemini-3-Pro, resulting in a rich, robust dataset. In contrast, online RL requires massive real-time interactions, restricting us to a single, comparatively weaker open-weight simulator (Qwen3-235B-A22B-Instruct-2507) due to latency and cost constraints. For the already capable SFT model, interacting exclusively with this narrower environment acts as a negative regularizer, causing it to overfit to the simulator's specific quirks and forget the generalized capabilities acquired during SFT.

\begin{table}[t]
  \centering
  \caption{Left: Performance comparison ($pass^1$) of different training paradigms using the CoVe framework on the Qwen3-4B-Instruct-2507 base model. \textbf{Bold} denotes the best results. Right: Success rates of SFT data generation across different domains using various User Simulator LLMs.}
  \vspace{3pt}
  \begin{minipage}{0.48\linewidth}
      \centering
      \scriptsize
      \renewcommand{\arraystretch}{1.3}
      \begin{tabular}{l|ccc}
          \toprule
          \textbf{Training Set} &
          \textbf{Airline} &
          \textbf{Retail} &
          \textbf{Average} \\
          \midrule
          Qwen3-4B-Inst. & 25.0\% & 40.1\% & 32.6\% \\
          SFT & \textbf{43.0\%} & \textbf{59.4\%} & \textbf{51.2\%} \\
          RL & 32.5\% & 48.9\% & 40.7\% \\
          SFT+RL & 38.0\% & 55.7\% & 46.9\% \\
          \bottomrule
      \end{tabular}
  \end{minipage}
  \hfill
  \begin{minipage}{0.48\linewidth}
      \centering
      \scriptsize
      \renewcommand{\arraystretch}{1.3}
      \begin{tabular}{l|ccc}
          \toprule
          \textbf{User Simulator} &
          \textbf{Airline} &
          \textbf{Retail} &
          \textbf{Average} \\
          \midrule
          Qwen3-235B-A22B-Inst. & 27.2\% & 50.2\% & 38.7\% \\
          Gemini-3-Pro & 62.8\% & 85.1\% & 74.0\% \\
          \bottomrule
      \end{tabular}
  \end{minipage}
  \label{tab:rl_sft_comparison_and_generation_success_rate}
  \end{table}

\paragraph{Dynamics of the Synthesis Process.}
Finally, we examine the trajectory-generation success rates of different User-Simulator LLMs across domains, as reported on the right side of Table~\ref{tab:rl_sft_comparison_and_generation_success_rate}, using Qwen3-235B-A22B-Instruct-2507 and Gemini-3-Pro as representative examples. The success rates vary substantially with the simulator’s capability. Gemini-3-Pro attains a higher average yield (74.0\%), largely because its advanced instruction-following ability allows it to accurately time the dialogue termination signal (e.g., \texttt{\#\#\#STOP\#\#\#}). In contrast, weaker simulators often misinterpret an agent's clarification request as task completion and prematurely halt the conversation, resulting in failed trajectories (this may also partially explain the suboptimal RL performance). Despite this variance in yield, we intentionally utilize a diverse ensemble of multiple models to synthesize our final SFT dataset. This diversity is crucial to emulate the varied conversational styles of real-world users and align with the rich evaluation scenarios of the $\tau^2$-bench. Additionally, the generation yield in the Retail domain consistently surpasses that of the Airline domain across all simulators. This trend is consistent with the comparatively poorer Airline results noted in our main evaluations, primarily due to the inherently higher difficulty of tasks within the Airline domain.
\section{Limitations and Future Work}
\label{sec:limit_future}

Currently, the sequential SFT+RL pipeline underperforms pure SFT due to the capability gap of the open-weight simulator used during online interaction. To unlock the full potential of RL and surpass pure SFT performance, future work will explore three directions: 1) adopting more capable general models as the User Simulator LLM, 2) training a dedicated User Simulator model by leveraging the context data collected from our existing generation pipeline, and 3) refining the existing simulator through prompt engineering to help it accurately time dialogue termination signals. 

Furthermore, our current study focuses strictly on the Airline and Retail domains of the $\tau^2$-bench. To comprehensively validate the framework's generalization capabilities across diverse and complex scenarios, future research will adapt the CoVe framework to additional domains, such as the Telecom domain within the $\tau^2$-bench, as well as other multi-turn interactive tool-use benchmarks like BFCL \citep{patil2025berkeley}.
\section{Conclusion}
\label{sec:conclusion}

In this paper, we introduce CoVe (Constraint-Verification), a novel post-training data synthesis framework designed to tackle the fundamental challenges of developing multi-turn interactive tool-use agents. By anchoring the generation process in explicit task constraints, CoVe successfully bridges the gap between the ambiguous nature of real-world user requests and the deterministic execution requirements of tools. Our framework employs constraint fuzzification to elicit complex, realistic multi-turn dialogues, while leveraging the original constraints as a deterministic checklist to ensure the absolute correctness of trajectory verification. Extensive evaluations on the highly challenging $\tau^2$-bench demonstrate CoVe's exceptional effectiveness and remarkable data efficiency. Notably, our compact CoVe-4B model significantly outperforms its base model and achieves performance competitive with open-source models up to $17\times$ its size (\eg, xLAM-2-70b-fc-r). By open-sourcing our trained models and the highly curated 12K interactive trajectory dataset, we aim to provide the research community with a robust, scalable foundation for advancing the next generation of reliable and capable interactive AI agents.



\bibliographystyle{plainnat}
\bibliography{references}






\end{document}